\documentclass{article}
\usepackage{spconf,amsmath,graphicx}

\usepackage{enumitem}
\setlist{nosep, leftmargin=14pt}

\usepackage{mwe} 
\usepackage{xcolor}
\usepackage{tabularx}
\usepackage{booktabs}
\usepackage{colortbl}
\usepackage{amsfonts}
\usepackage{url}
\usepackage{pifont}

\newcommand{\dino}{\texttt{DINOv2}~\cite{oquab2023dinov2} }
\newcommand{\clipdnn}{\texttt{CLIP}~\cite{radford2021learning} }
\newcommand{\gdino}{\texttt{G-DINO}~\cite{liu2024grounding} }
\newcommand{\cmark}{\textcolor{blue}{\ding{51}}} 
\newcommand{\xmark}{\textcolor{red}{\ding{55}}}   

\def\vit{\texttt{ViT}\xspace}
\def\vits{\texttt{ViTs}\xspace}
\def\bcdm{\texttt{BCDM}\xspace}
\def\rois{\texttt{RoIs}\xspace}
\def\roi{\texttt{RoI}\xspace}
\def\bce{\texttt{BCE}\xspace}
\def\mlp{\texttt{MLP}\xspace}
\def\auc{\texttt{AUC}\xspace}
\def\sota{\texttt{SOTA}\xspace}
\def\mammoclip{\texttt{Mammo-CLIP}\xspace}
\def\vindr{\texttt{VinDR}\xspace}
\def\mmbcd{\texttt{MMBCD}\xspace}
\newcommand{\myfirstpara}[1]{\noindent \textbf{#1.}~}
\newcommand{\mypara}[1]{\vspace{0.2em}\myfirstpara{#1}}

\title{Attend what matters: Leveraging vision foundational models for breast cancer classification using mammograms}

\name{
\shortstack[c]{
Samyak Sanghvi$^{\dagger}$\quad Piyush Miglani$^{\ddagger}$\quad Sarvesh Shashikumar$^{\dagger}$ \\
Kaustubh R Borgavi$^{\ddagger}$\quad Veenu Singla$^{\S}$\quad Chetan Arora$^{\dagger\ddagger}$
}
}

\address{
\{$^{\dagger}$ Department of Computer Science and Engineering, $^{\ddagger}$ Yardi School of AI\}, IIT Delhi, New Delhi, India \\
$^{\S}$ Department of Radiodiagnosis, PGIMER Chandigarh, Chandigarh, India
}

\begin{document}
%
\maketitle

\begin{abstract}

Vision Transformers (\vits) have become the architecture of choice for many computer vision tasks, yet their performance in computer-aided diagnostics remains limited. Focusing on breast cancer detection from mammograms (\bcdm), we identify two main causes for this shortfall. First, medical images are high-resolution with small abnormalities, leading to an excessive number of tokens and making it difficult for the softmax-based attention to localize and attend to relevant regions. Second, medical image classification is inherently fine-grained, with low inter-class and high intra-class variability, where standard cross-entropy training is insufficient. To overcome these challenges, we propose a framework with three key components: (1) Region of interest (\roi) based token reduction using an object detection model to guide attention; (2) contrastive learning between selected \rois to enhance fine-grained discrimination through hard-negative based training; and (3) a \dino pretrained \vit that captures localization-aware, fine-grained features instead of global \clipdnn representations. Experiments on public mammography datasets demonstrate that our method achieves superior performance over existing baselines, establishing its effectiveness and potential clinical utility for large-scale breast cancer screening. Our code is available for reproducibility here: 
\newline https://aih-iitd.github.io/publications/attend-what-matters

\end{abstract}
\begin{keywords}
Mammography, Classification, Contrastive Learning
\end{keywords}
\section{Introduction}
\label{sec:intro}

\myfirstpara{Background} 
Mammography is a low-dose X-ray imaging modality that serves as the most widely adopted screening procedure for the early detection of breast cancer \cite{rakhunde2022thermography}, the most common malignancy among women, accounting for more than 685,000 deaths worldwide in 2020 \cite{lei2021global}. As the gold standard for detecting breast malignancies \cite{rakhunde2022thermography}, mammograms provide radiologists with information-rich, high-resolution $12$-$16$ bit representations of each breast at 4K$\times$4K resolution from two views: the Mediolateral Oblique (MLO) or side view, and the Craniocaudal (CC) or top view. This facilitates the identification and detection of suspicious masses and lesions within the breast tissue, which are early indicators of the presence of breast cancer in the patient.


\mypara{Challenges} 
%
The potential of AI models to achieve expert-level diagnostic performance has driven research in applications of Deep Neural Network-based breast cancer screening via mammography \cite{alberdi2005use, dang2022impact, laang2021identifying, larsen2022possible, sharma2023multi}. Extensive prior work have explored Convolutional Neural Network and Transformer-based models, demonstrating promising, yet sub-par results in the real world setting. A prevalent practice among the transformer based methods involves downsampling mammograms to resolutions such as 1024 $\times$ 1024 or lower to reduce computational overhead. Furthermore, medical imaging is to be treated as a fine-grained classification problem, specifically in case of mammograms, where the cancerous region occupies around 0.1\% of the entire image \cite{zhao2025boltzmann}. Using the original high-resolution mammography images results in a substantially increased number of input tokens for Transformer models. This high token count, when processed through the softmax-normalized attention mechanism, defocus the attention distribution, making it harder for the model to effectively attend to the sparse and diagnostically relevant regions, and increasing the risk of misclassification.

\mypara{Our solution} 
In this work, we address these challenges by proposing a novel training paradigm that first identifies \rois from the full-resolution mammogram using an object detection model, specifically \gdino. Instead of processing entire mammograms or their aggressively downsampled versions, we focus exclusively on the \rois predicted by the detection module in response to clinically relevant text prompts. Since these \rois lack explicit ground truth labels, training solely based on the object detector’s outputs is insufficient. To extract meaningful features from each \rois crop, we employ \dino to generate discriminative embeddings. We further introduce a novel loss that combines binary cross-entropy (\bce) with a contrastive loss applied between embeddings of the \rois, enabling the model to learn from hard negatives, given that typically only one \roi contains the tumor. By constraining learning to anatomically meaningful \rois and leveraging the evolving capabilities of \dino, our approach preserves critical diagnostic details while maintaining adaptability to newer model versions. Additionally, we propose an attention-based aggregation mechanism that emphasizes diagnostically relevant \rois, ensuring sensitivity to subtle, localized features most indicative of malignancy. 

\mypara{Contributions}
Our key contributions are as follows:

\begin{enumerate}[leftmargin=*]
    \item We employ an object detection module as a preprocessor for full scale mammograms in order to obtain \rois, which result in fewer tokens that need to be attended to by the classification head.
    \item To improve fine-grained classification, we use contrastive training between the selected \rois. This leverages the insight that medical abnormalities are typically localized within a single \roi, allowing the model to learn from hard negatives effectively.
    \item Instead of using a \clipdnn pretrained Vision Transformer (\vit) trained for global feature extraction, we adopt a \dino pretrained \vit, which is trained on multiple localization tasks to extract fine-grained, local features.
	\item Comparison of baselines on public mammography datasets establish the framework's efficiency, adaptability, and clinical utility for large-scale breast cancer screening applications. Our proposed approach achieves an increment of \textbf{1\%} on AUC and a remarkable \textbf{4\%} gain over previously reported state-of-the-art classifier, which requires an image-text pretraining on large private datasets, unlike our simplified image-only training with publicly available training datasets, making it more accessible.
\end{enumerate}

\section{Methodology}
\label{method}

\begin{figure*}[t] 
	\centering
	\includegraphics[width=\linewidth, height=0.30\textheight]{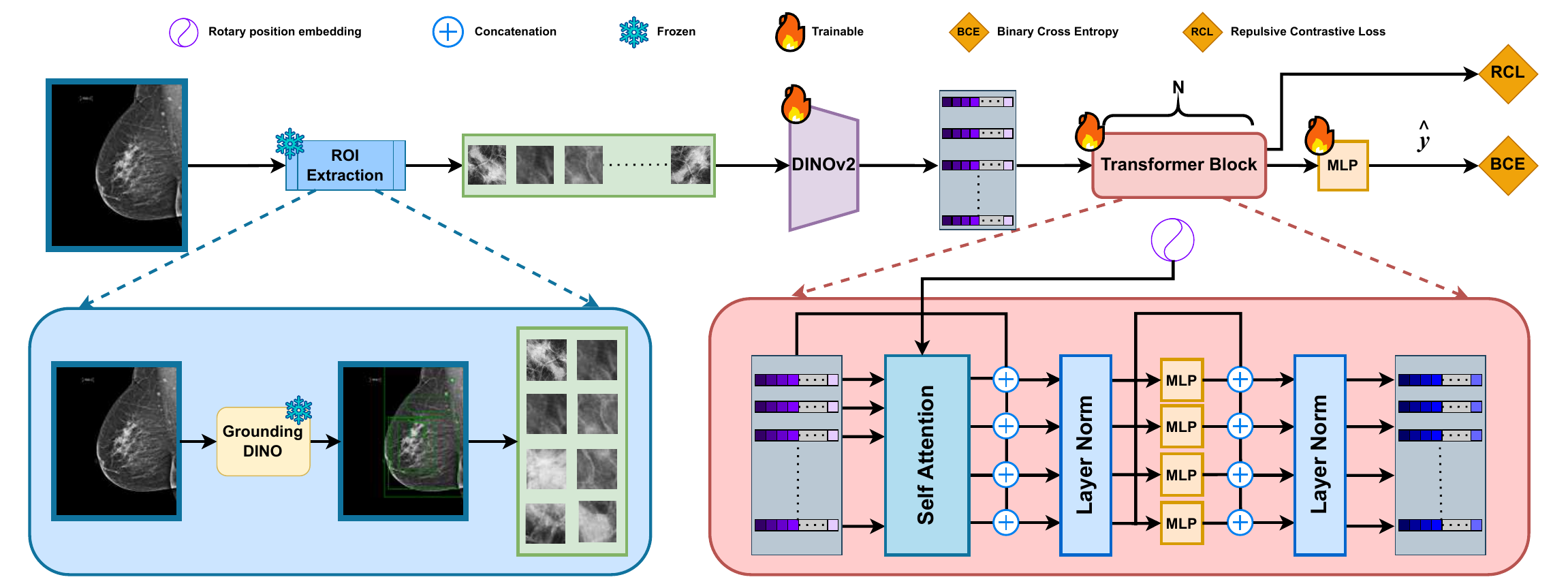} 
	\caption{\textbf{Overview.} In our proposed method, we first use G-DINO in a zero-shot setting to localize semantically important Regions of Interest (RoIs) based on a clinically meaningful text prompt. The top-$k$ RoIs, selected by bounding box confidence scores, are cropped and passed through a pretrained DINOv2 encoder to extract $k$ feature vectors. These features are processed by a transformer module, enabling interaction among regions through self-attention. The global RoI (covering the full breast) is used as the final representation which is contextually enriched by fine-grained lesion-level RoIs, and is fed into a Multi-Layer Perceptron (MLP) for classification.}

	\label{fig:architecture}
\end{figure*}

\myfirstpara{Problem Statement}
Our goal in this study is to develop a classification framework $f(x)$ that maps 2D mammograms ($x \in \mathcal{X}$) to binary labels indicating the presence or absence of breast cancer, where $\mathcal{X}$ denotes a mammography dataset.

\mypara{Attention between \rois}
As shown in Figure~\ref{fig:architecture}, we extract the \texttt{RoI} using \texttt{G-DINO}~\cite{liu2024grounding} and subsequently employ \texttt{DINOv2}~\cite{oquab2023dinov2} for feature extraction. Detailed implementation details for these two initial stages are provided in the supplementary material on our project page. After feature extraction, we obtain $k$ fixed-length feature vectors $\{ \mathbf{z}_i \in \mathbb{R}^{a} \}_{i=1}^k$, where each $\mathbf{z}_i$ represents the \dino-encoded embedding of the $i$-th cropped \rois. Mammograph pathologies are typically accompanied by associated architectural distortions in the nearby regions. Hence, to capture the spatial and relational correlation between these regions, we make use of self-attention and layer normalization \cite{vaswani2017attention}, and additionally employ \textbf{Rotary Position Embeddings (RoPE)}~\cite{su2024roformer}, which inject positional information by rotating query and key representations in the attention mechanism, rather than through explicit addition or concatenation. This rotation implicitly encodes relative position information within the dot-product attention, allowing the model to reason about spatial relationships among RoIs without any additive positional embeddings, while facilitating interactions between fine-grained and coarse-grained regions across the mammogram.
Thus, the input to the transformer is simply:
\[
\mathbf{X}_0 = \{ \mathbf{z}_1, \dots, \mathbf{z}_k \}.
\]
This sequence is passed through a stack of transformer blocks for attention-based contextualization.

\mypara{Attend Where It Matters}
In our design, attention is restricted to only those regions that \gdino identified as clinically significant, effectively filtering out irrelevant background and noise, and subsequently reducing the input tokens required for the task. Inclusion of \dino as the feature encoder ensures that all attention is concentrated on regions with diagnostic potential, including subtle abnormalities that may otherwise be overlooked in a global processing approach.
After the final transformer layer, we obtain $\mathbf{X}_L = \{ \mathbf{x}_1', \dots, \mathbf{x}_k' \}$. To produce a global breast-level representation, we select $\mathbf{x}_1'$ which corresponds to the full-breast \texttt{RoI} (the largest bounding box), which has attended to all $k-1$ fine-grained RoIs encoding local patterns such as micro-calcifications or masses, or other visual cues that indicate the presence of a malignant tumor. The representation, now enriched with localized cues from all attended \rois, is processed through an \mlp block to yield a classification score that reflects both semantic richness and global contextual awareness.

\mypara{Loss Function}
%
%
We use the standard Binary Cross-Entropy (BCE) loss between the final predicted probability $\hat{y}$ and the ground truth label $y \in \{0, 1\}$, defined as:
\begin{equation}
    \mathcal{L}_{\text{BCE}} = -\left[ y \cdot \log(\hat{y}) + (1 - y) \cdot \log(1 - \hat{y}) \right]
    \label{eq:bce_loss}
\end{equation}
To encourage feature diversity among different \roi proposals, we introduce a repulsive contrastive loss that penalizes high similarity between distinct embeddings. This loss computes normalized pairwise cosine similarity between the \roi embeddings-that are the result of the final transformer block, and minimizes the squared off-diagonal elements, effectively pushing different proposals to occupy distinct regions in the embedding space. We exclude the anchor or $\mathbf{x}_1'$ (full-breast) embedding and applying this repulsion only to the other fine-grained \rois, the loss prevents feature collapse while maintaining computational efficiency. The repulsive term complements the \bce loss by enforcing both correct classification and embedding diversity, leading to more discriminative representations across different anatomical regions. The overall loss function can be written as:
\begin{equation}
    \mathcal{L}_{\text{rep}} = \frac{1}{B} \sum_{b=1}^{B} \frac{1}{K(K-1)} \sum_{i \neq j} \text{sim}(\mathbf{e}_i^{(b)}, \mathbf{e}_j^{(b)})^2
    \label{Repulsive contrastive loss}
\end{equation}
where \(B\) is the batch size, \(K\) is the number of \roi proposals, \(\mathbf{e}_i^{(b)}\) is the normalized embedding of the \(i\)-th \roi in batch \(b\), and \(\text{sim}(\cdot, \cdot)\) denotes cosine similarity.

\definecolor{lightBlue}{RGB}{225, 240, 255}
\definecolor{blue}{RGB}{0, 102, 204}
\definecolor{mygray}{gray}{0.85}

\begin{table}[t]
	\centering
	\setlength{\tabcolsep}{4pt}
	\begin{tabular}{lccccc}
		\arrayrulecolor{black}\specialrule{0.1em}{0em}{0em}
		\rowcolor{mygray}
		\textbf{Method} & \textbf{AUC} & \textbf{F1} & \textbf{R@0.1} & \textbf{R@0.3} & \textbf{R@0.5} 
		\\
		\arrayrulecolor{black}\specialrule{0.1em}{0em}{0em}
		\rowcolor{lightBlue}
		\multicolumn{6}{c}{\textit{Vision only}} 
		\\
		ViT-A \cite{huix2024natural} & 79.0 & 41.1 & 55.0 & 71.2 & 84.3 \\
		ViT-B \cite{huix2024natural} & 83.0 & 50.0 & 61.4 & 77.0 & 86.9 \\
		ViT-C \cite{huix2024natural} & 78.4 & 31.1 & 43.7 & 67.2 & 82.4 \\
		MedVAE\cite{varma2025medvae} &57.5 &20.6 &23.7 &41.9 &60.1 \\
        TReg-SwinT\cite{nguyen2023transreg} & 85.8 & 53.0 & 55.1 & 80.6 & 90.2 \\
		XFMamba\cite{zheng2025xfmamba} & 63.6 & 18.3 & 25.2 & 51.5 & 64.6 \\
		\arrayrulecolor{black}\specialrule{0.1em}{0em}{0em}
		\rowcolor{lightBlue}
		\multicolumn{6}{c}{\textit{Image-Text}} 
		\\
		MMBCD\cite{jain2024mmbcd} & 77.1 & 27 & 50 & 66.2 & 82.8 
		\\
		M-C-B5 \cite{ghosh2024mammo} & 85.8 & 50.8 & 65.4  & \textbf{83.5} & 89.9 
		\\
		\rowcolor{mygray}
		\textbf{Ours} & \textbf{86.6} & \textbf{54.5} & \textbf{66.5} & 80.7 & \textbf{90.3} 
		\\
		\arrayrulecolor{black}\specialrule{0.1em}{0em}{0em}
	\end{tabular}
	\caption{Performance comparison on the \vindr dataset. ViT-A and ViT-B correspond to \dino with a linear layer at input resolutions of 448$\times$448 and 1024$\times$1024 respectively. ViT-C demonstrates scores using a \texttt{DeIT} head instead of a linear layer at 448$\times$448 input resolution.}
	\label{tab:main_results}
\end{table}

\begin{table}[ht]
\centering
\setlength{\tabcolsep}{4pt} 
\footnotesize
\begin{tabular}{ccccc|c|c}
\toprule
\dino & \gdino & Anchor & RoPE & RCL & AUC & F1 \\
\midrule
\cmark & \xmark & \xmark & \xmark & \xmark & 83.0 & 50.0 \\
\cmark & \cmark & \xmark & \xmark & \xmark & 76.4 & 49.9 \\
\cmark & \cmark & \cmark & \xmark & \xmark & 83.9 & 51.9\\
\cmark & \cmark & \cmark & \cmark & \xmark & 84.4 & 53.8 \\
\cmark & \cmark & \cmark & \cmark & \cmark & \textbf{86.5} & \textbf{54.4} \\
\bottomrule
\end{tabular}
\caption{Experimental results for different combinations of utilizing \dino embeddings, \gdino RoI crops, RoPE and RCL.}
\label{tab:vindr_ablations}
\end{table}

\section{Results}
\label{experiments}

In Table \ref{tab:main_results}, we compare our results with those of other unimodal image-based and multimodal image-text-based baselines, with the previous state-of-the-art (\sota) being \mammoclip \cite{ghosh2024mammo}. Despite being a unimodal approach, our scores demonstrate a \textbf{4\%} improvement on the F1 score and a \textbf{1\%} gain on the \auc over this \sota, which relies on \clipdnn-style pretraining requiring text supervision, and use of training on private dataset, which makes the technique inaccessible for reproducing. While \mammoclip can be inferred without text after fine-tuning, the subsequent baseline \mmbcd \cite{jain2024mmbcd} requires text at runtime. We fine-tune \mmbcd \cite{jain2024mmbcd} using attributes from \vindr \cite{nguyen2023vindr} as training text. However, during inference, we pass an empty string for textual input because \vindr lacks unique findings for non-malignant cases, avoiding bias learned by the model.

Experiments on \texttt{ViTs A, B, C} explore different fine-tuning settings of \dino with varying classification heads and input sizes. Image-only baselines like \texttt{MedVAE} \cite{varma2025medvae}, \texttt{XFMamba} \cite{zheng2025xfmamba}, and \texttt{TransReg} \cite{nguyen2023transreg} perform notably lower. All scores here are computed under the binary classification setting, without averaging across classes.

Table \ref{tab:vindr_ablations} presents ablation study to evaluate the impact of each component. Starting with a fine-tuned \dino with a linear classification layer achieved an AUC of 83\% and an F1 score of 50\%, setting a baseline with mammograms resized to $1024 \times 1024$. Introducing \gdino to extract \rois avoids resizing-related information loss. We max-pooled embeddings of all \rois passed through an \mlp and trained using only \bce loss. This performed similarly proving the efficiency of not requiring to pass all input tokens.

Passing only the anchor embedding—the \roi covering the entire breast after attending to other \rois within the Transformer self-attention block \cite{vaswani2017attention}—to the final \mlp improved the F1 score by nearly 2\%. Incorporating \texttt{RoPE} \cite{su2024roformer} to encode spatial information of non-consecutive, non-adjacent \rois further increased the F1 score by almost \textbf{2\%}. Lastly, adding a repulsive contrastive loss to separate dissimilar \rois boosted both the F1 score and \auc by \textbf{1\%}, achieving our best performance.

\section{Conclusion}

In this study, investigated the lower performance of transformer models in medical imaging, and came up with large number of tokens due to high resolution, and fine-grained nature of the problem as the reasons. We presented a novel architecture based on \roi based token selection, contrastive loss based hard negative training, and upgraded \vit backbone to successfuly overcome the challenges.
\section{Compliance with Ethical Standards}
This research was conducted retrospectively using human subject data obtained from an open-access source \cite{nguyen2023vindr}. Ethical approval was not required as confirmed by the license attached with the open access data.
\label{sec:compliance_statement}

\bibliographystyle{IEEEbib}
\bibliography{strings,refs}

@article{rakhunde2022thermography,
  title={Thermography as a breast cancer screening technique: A review article},
  author={Rakhunde, Manasi B and others},
  journal={Cureus},
  volume={14},
  number={11},
  year={2022},
  publisher={Cureus}
}

@article{lei2021global,
  title={Global patterns of breast cancer incidence and mortality: A population-based cancer registry data analysis from 2000 to 2020},
  author={Lei, Shaoyuan and others},
  journal={Cancer Communications},
  volume={41},
  number={11},
  pages={1183--1194},
  year={2021},
  publisher={Wiley Online Library}
}

@article{nguyen2023transreg,
  title={TransReg: Cross-transformer as auto-registration module for multi-view mammogram mass detection},
  author={Nguyen, Hoang C and others},
  journal={arXiv preprint arXiv:2311.05192},
  year={2023}
}

@inproceedings{radford2021learning,
  title={Learning transferable visual models from natural language supervision},
  author={Radford, Alec and others},
  booktitle={ICML},
  year={2021},
}

@inproceedings{jain2024mmbcd,
  title={MMBCD: multimodal breast cancer detection from mammograms with clinical history},
  author={Jain, Kshitiz and others},
  booktitle={MICCAI},
  year={2024},
}

@inproceedings{ghosh2024mammo,
  title={Mammo-clip: A vision language foundation model to enhance data efficiency and robustness in mammography},
  author={Ghosh, Shantanu and others},
  booktitle={MICCAI},
  year={2024},
}

@article{zheng2025xfmamba,
  title={XFMamba: Cross-Fusion Mamba for Multi-View Medical Image Classification},
  author={Zheng, Xiaoyu and others},
  journal={arXiv preprint arXiv:2503.02619},
  year={2025}
}

@article{varma2025medvae,
  title={MedVAE: Efficient automated interpretation of medical images with large-scale generalizable autoencoders},
  author={Varma et al., Maya},
  journal={arXiv},
  year={2025}
}

@inproceedings{huix2024natural,
  title={Are natural domain foundation models useful for medical image classification?},
  author={Huix, Joana Pal{\'e}s and others},
  booktitle={WACV},
  year={2024}
}

@article{oquab2023dinov2,
  title={Dinov2: Learning robust visual features without supervision},
  author={Oquab, Maxime and others},
  journal={arXiv preprint arXiv:2304.07193},
  year={2023}
}

@article{vaswani2017attention,
  title={Attention is all you need},
  author={Vaswani, Ashish and others},
  journal={NeurIPS},
  year={2017}
}

@article{nguyen2023vindr,
  title={VinDr-Mammo: A large-scale benchmark dataset for computer-aided diagnosis in full-field digital mammography},
  author={Nguyen, Hieu T and others},
  journal={Scientific Data},
  year={2023},
}

@article{alberdi2005use,
  title={Use of computer-aided detection (CAD) tools in screening mammography: a multidisciplinary investigation},
  author={Alberdi, Eugenio and others},
  journal={The British journal of radiology},
  year={2005},
}

@inproceedings{zhao2025boltzmann,
  title={Boltzmann Attention Sampling for Image Analysis with Small Objects},
  author={Zhao, Theodore and others},
  booktitle={CVPR},
  year={2025}
}

@article{dang2022impact,
  title={Impact of artificial intelligence in breast cancer screening with mammography},
  author={Dang, Lan-Anh and others},
  journal={Breast Cancer},
  year={2022},
}

@article{laang2021identifying,
  title={Identifying normal mammograms in a large screening population using artificial intelligence},
  author={L{\aa}ng, Kristina and others},
  journal={European radiology},
  year={2021},
}

@article{larsen2022possible,
  title={Possible strategies for use of artificial intelligence in screen-reading of mammograms, based on retrospective data from 122,969 screening examinations},
  author={Larsen, Marthe and others},
  journal={European radiology},
  year={2022},
}

@article{sharma2023multi,
  title={Multi-vendor evaluation of artificial intelligence as an independent reader for double reading in breast cancer screening on 275,900 mammograms},
  author={Sharma, Nisha and others},
  journal={BMC cancer},
  year={2023},
}

@inproceedings{liu2024grounding,
  title={Grounding dino: Marrying dino with grounded pre-training for open-set object detection},
  author={Liu, Shilong and others},
  booktitle={ECCV},
  year={2024},
}

@article{su2024roformer,
  title={Roformer: Enhanced transformer with rotary position embedding},
  author={Su, Jianlin and others},
  journal={Neurocomputing},
  year={2024},
}

\end{document}